\documentclass[letterpaper, 10 pt, conference]{ieeeconf}  

\IEEEoverridecommandlockouts                              

\usepackage{graphicx} 
\usepackage{amsmath} 
\usepackage{amssymb}  
\usepackage[caption=false,font=footnotesize]{subfig}
\usepackage{cite}
\makeatletter
\let\cl@chapter\undefined
\makeatother
\usepackage{cleveref}
\usepackage{flushend} 
\usepackage{cite}

\title{\LARGE \bf
Path-Following Control and Terramechanics Analysis for \\Planetary Rovers Under Wheel-to-Wheel Traction Asymmetry
}

\author{Ryuya Matsuoka$^{1}$, Keisuke Takehana$^{2}$, Kentaro Uno$^{1}$, Toshinori Kuwahara$^{2,1,3}$ and Kazuya Yoshida$^{1}$
\thanks{*This work was supported by JSPS KAKENHI, Grant Number 24KJ0413.}
\thanks{$^{1}$R. Matsuoka, K. Uno, T. Kuwahara, and K. Yoshida are with the Department of Aerospace Engineering, Graduate School of Engineering, Tohoku University, Sendai, Miyagi 980-8579, Japan. Email: \texttt{\small matsuoka.ryuya.p4@dc.tohoku.ac.jp}.}
\thanks{$^{2}$K. Takehana and T. Kuwahara are with the Research Center for Green X-Tech, Green Goals Initiative, Tohoku University. Email: \texttt{\small toshinori.kuwahara.b3@tohoku.ac.jp}.}
\thanks{$^{3}$T. Kuwahara is with the Research Center for Space Cross-Tech, Green Goals Initiative, Tohoku University.}
}

\begin{document}

\maketitle
\thispagestyle{empty}
\pagestyle{empty}

\begin{abstract}
This paper proposes a control strategy for path following that is model-free and relies solely on deceleration for skid-steering planetary rovers navigating deformable loose terrain under continuously imposed traction asymmetry. While conventional controllers that are based on kinematics frequently cause slip-sinkage entrapment by accelerating the wheels during path correction, the proposed approach prevents this failure by setting an upper limit on the maximum commanded velocity. Heading correction is achieved solely through the selective deceleration of the outer wheels, which are located on the outside of the turn, driving them into a negative slip regime to act as a mechanical anchor. The system was evaluated using the four-wheel independent-drive rover EX1 under an asymmetric wheel configuration with different left and right grouser heights that induces significant deviations from the path. Experimental results demonstrate that this deceleration-only control successfully suppresses accumulated lateral drift across various velocity regimes up to $0.7\,\mathrm{m/s}$ without causing sinkage. Crucially, direct force measurements from onboard multi-axis sensors provide important empirical evidence of the underlying terramechanics, proving that the targeted deceleration establishes dynamic load equalization across the chassis and completely restores the native thrust capability of the opposite driving wheel.
\end{abstract}


\section{Introduction}
Recent lunar exploration initiatives, such as the Artemis program \cite{smith2020artemis}, are shifting planetary exploration from localized observations at low speeds to operations over wide areas at high speeds. To achieve these objectives, the next generation of rovers must traverse rough terrains over long distances at higher speeds. This shift in mobility requirements demands a reevaluation of conventional locomotion and control strategies historically designed for platforms that move slowly.

Unlike terrestrial vehicles operating on rigid roads, planetary rovers travel on deformable terrain composed of loose regolith. Consequently, rover mobility is governed by wheel--soil interaction. The generation of traction, sinkage, motion resistance, and wheel slip are all strongly affected by terrain deformation, making mobility prediction and control significantly more challenging than in conventional ground vehicles. 

\begin{figure}[t]
    \centering
    \includegraphics[width=1.0\linewidth]{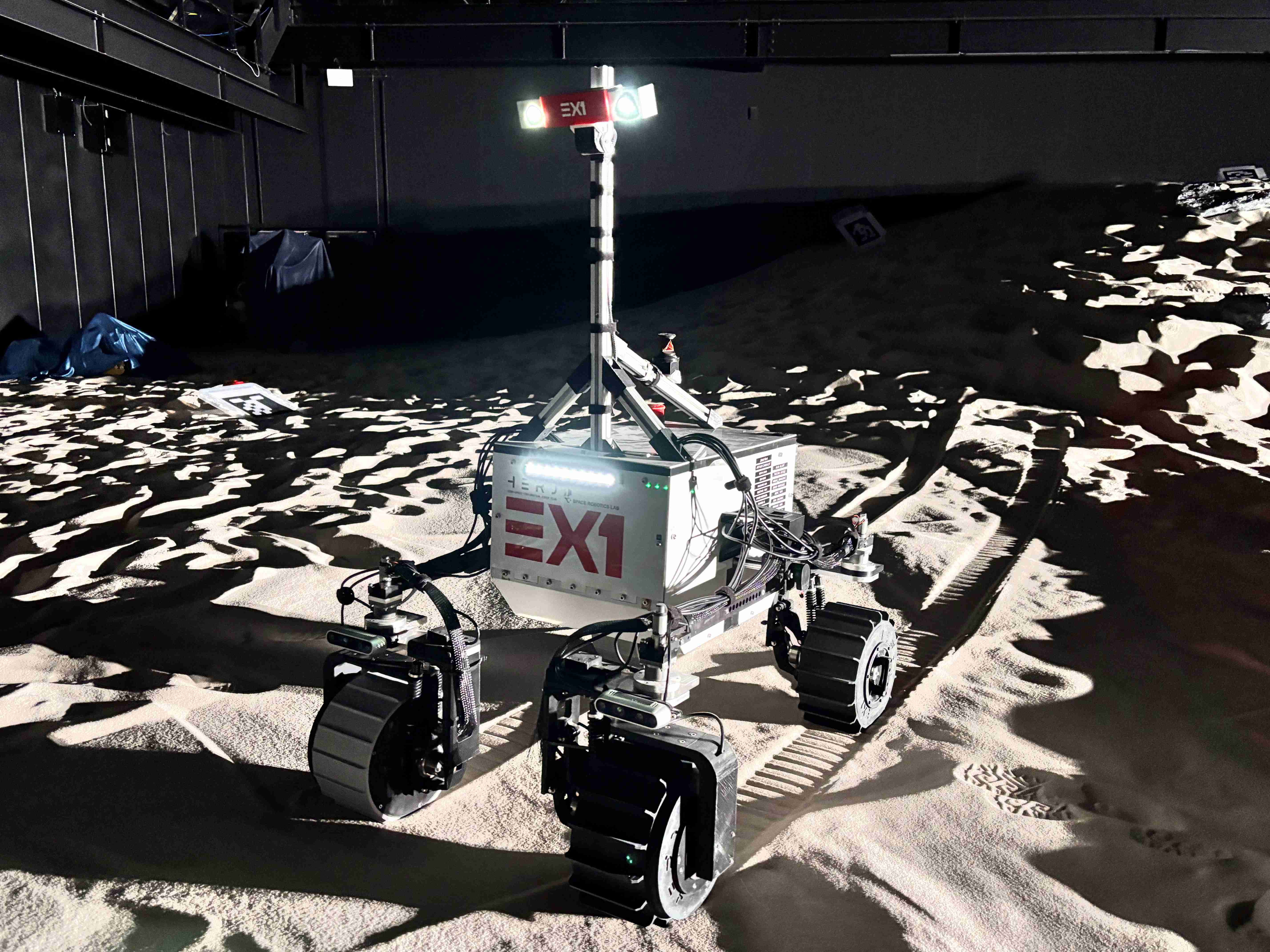}
\caption{Overview of the experimental rover EX1 used in this study. The rover employs a four-wheel configuration and is equipped with onboard force/torque sensors for measuring wheel-terrain interaction during locomotion on loose soil.}
    \label{fig:ex1_main}
\end{figure}

\subsection{Related Works}

Traversing the lunar surface presents fundamental terramechanics challenges arising from complex wheel--soil interactions. Fundamental terramechanics theories dictate that wheel slip inevitably increases mechanical sinkage and locomotion resistance through soil shear failure \cite{bekker1,bekker2,bekker3,wong2022theory}. As documented by NASA's Jet Propulsion Laboratory (JPL), unpredicted wheel slip on loose sand fields leads to localization errors and vehicle immobilization, which motivated the development of visual odometry \cite{maimone2007two} and predictive slip-learning algorithms~\cite{angelova2006learning}. To mitigate these path deviations and associated hazards, various control frameworks for path following and slip compensation have been developed based on rover motion dynamics and terrain parameter estimation \cite{ishigami2006path,ishigami2009slope, iagnemma2004mobile}. Furthermore, strategies for torque vectoring and traction distribution have been proposed to stabilize vehicles that utilize skid steering on soft and sloping terrains \cite{liang2020torque, liu2025torque}. While recent hardware advancements attempt to optimize traction dynamically using variable grouser heights that adapt to the terrain \cite{griffo2026terrain}, these active modifications significantly increase system complexity and mass. Consequently, control strategies based on software remain crucial.

However, conventional control approaches often share two major limitations when applied to exploration at high speeds on loose ground: (i) they rely on the estimation of uncertain soil parameters (e.g., cohesion, internal friction angle, and slip ratios) that exhibit nonlinear fluctuations, and (ii) they often command the acceleration of the outer wheels to compensate for heading drift. On loose sand, forcing a wheel to accelerate inevitably induces soil shear failure rather than generating forward thrust, which can be modeled through soil mechanics based on an equivalent rigid wheel \cite{shibly2005equivalent} and depends on the structural constraints of the wheel grousers \cite{Chris2012}. This slip triggers the slip-sinkage effect, where the wheel physically excavates the soil, increasing bulldozing resistance and subsequently leading to immobilization. This vulnerability has been demonstrated in past planetary missions; NASA's Spirit rover suffered permanent immobilization on a soft sand field because continuous wheel rotation under conditions of high slip triggered failure due to slip-sinkage \cite{callas2015mars}, while the Curiosity rover experienced severe slip trends on Martian dunes, which required mission operators to alter paths to avoid immobilization \cite{rankin2021mars}. Furthermore, experimental studies on rovers operating at high speeds have shown that excessive slip significantly degrades net traction efficiency and drawbar pull \cite{takehana2025_JTerra}. Consequently, the conventional paradigm of control that permits or commands wheel acceleration during path recovery inherently carries a risk of immobilization.

\subsection{Contributions}

Tracking paths on soft terrain thus requires a focus on preserving soil mobility over the strict tracking of kinematic velocity. To address this issue, this paper presents a model-free, deceleration-only control strategy designed as a conservative, fail-safe safety patch for skid-steering rovers. Modern state-of-the-art methods, such as Extended Kalman Filter (EKF)-based slip estimators or Visual Odometry (VO)-based traction controllers, can dynamically mitigate slip-sinkage; however, they rely heavily on accurate soil parameter identification and computationally intensive vision algorithms that remain vulnerable to featureless lunar regolith and sudden lighting variations. In contrast, our approach deliberately bypasses online soil estimation and wheel acceleration entirely. By capping the maximum commanded velocity at a predefined nominal reference speed ($v_{\mathrm{ref}}$), heading corrections are achieved solely through the selective deceleration of the outer wheels. This targeted deceleration drives the designated wheel into a negative slip regime, inherently utilizing terrain resistance to act as a mechanical anchor. Driven by the dynamics of the rigid-body chassis (i.e., the seesaw effect), this braking force is mirrored across the lever arm of the chassis, establishing a symmetric dissociation of bilateral traction that thrusts the opposite driving wheel forward \cite{yoshida2002motion}. While this constraint on deceleration reduces the overall traveling speed during active turning maneuvers, it prevents cycles of high slip and eliminates the risk of slip-sinkage. In planetary exploration, where immobilization leads to mission failure, prioritizing fail-safe mobility on soil over the strict maintenance of speed is a practical engineering requirement.

The main contributions of this paper are summarized as follows:
\begin{itemize}
    \item We propose a model-free control strategy based on deceleration that utilizes a constraint on velocity saturation to avoid the slip-sinkage cycle, thereby entirely eliminating the need for the complex estimation of slip online and the identification of soil parameters.
    \item We establish an experimental benchmark to validate the capability of the controller for path recovery by configuring the rover with an asymmetric arrangement of the wheels that features different grouser heights between the left and right sides, thereby inherently enforcing a steady imbalance of thrust on loose terrain.
    \item We conduct an analysis of dynamic traction using direct measurements of force obtained in real time from multi-axis sensors mounted onboard. This analysis provides empirical evidence of the underlying terramechanics, demonstrating how selective deceleration effectively equalizes dynamic normal loads and triggers a symmetric dissociation of traction to restore the capability for thrust without causing sinkage across various regimes of velocity up to $0.7\,\mathrm{m/s}$.
\end{itemize}

\section{Rover System and Kinematics}

\subsection{The EX1 Rover System}
The experimental rover EX1 \cite{rodriguez2023enabling}, shown in Fig.~\ref{fig:ex1_main}, is a platform that utilizes four-wheel independent drive and skid steering, designed for research on locomotion at high speeds on loose soil. As summarized in Table~\ref{tab:ex1_params}, the rover measures $820\,\mathrm{mm}$ in length and $520\,\mathrm{mm}$ in width, with a total mass of approximately $24\,\mathrm{kg}$. Each wheel is equipped with an in-wheel motor, which enables the velocity of individual wheels to be independently controlled. While conventional rovers for planetary exploration typically operate at low speeds, EX1 can achieve a maximum speed of $1.0\,\mathrm{m/s}$, effectively allowing for the investigation of the dynamics of locomotion at high speeds.

For the estimation of vehicle states and the acquisition of data, the rover is equipped with onboard sensors. A 9-axis Inertial Measurement Unit (IMU) mounted at the center of the chassis provides the yaw rate ($\omega_z$) in real time for feedback control. Additionally, a 6-axis force and torque sensor is integrated into each of the wheel assemblies. This configuration inherently enables the direct measurement of dynamic drawbar pull ($F_x$) and vertical load ($F_z$) acting on individual wheels during locomotion, thereby providing direct physical data regarding the interactions between the wheels and the terrain.

\subsection{Asymmetric wheel configuration}

\begin{figure}[t]
    \centering
    \includegraphics[width=1.0\linewidth]{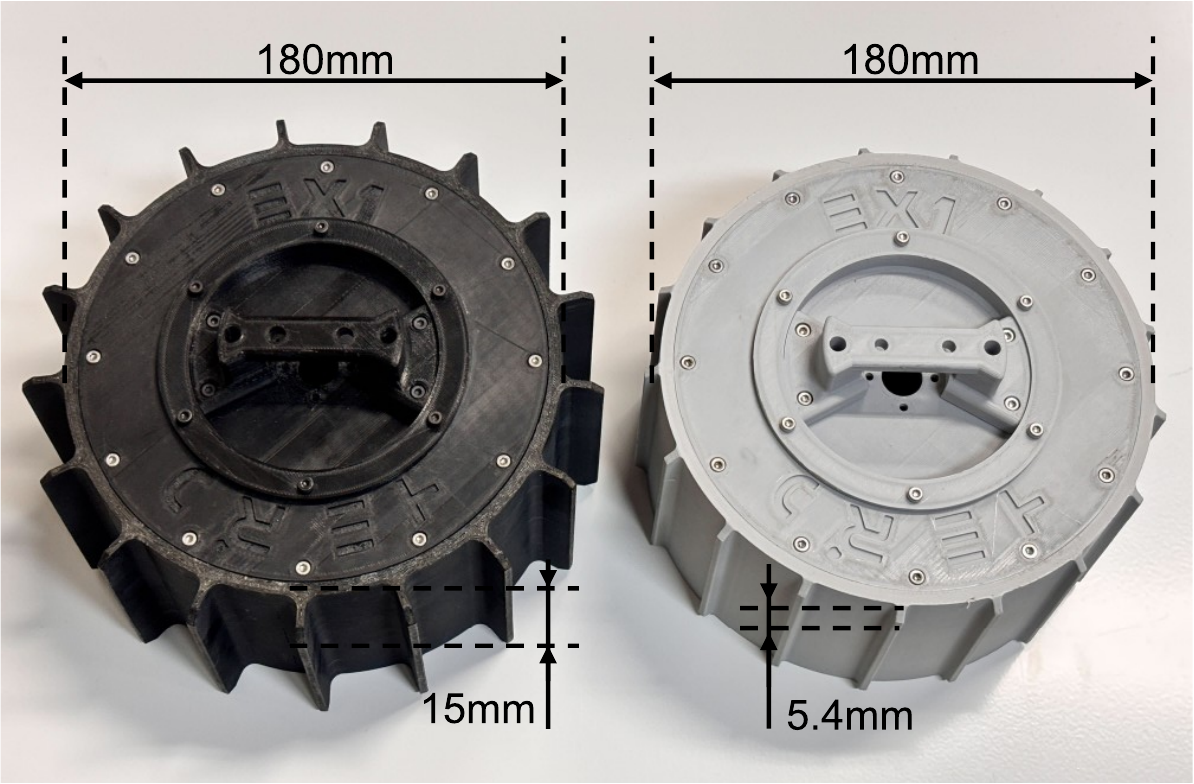}
\caption{Asymmetric wheel configuration used to generate a sustained traction imbalance. Different grouser heights (15.0 mm on the left and 5.4 mm on the right) produce unequal wheel-soil interaction forces, resulting in a persistent yaw disturbance.}
    \label{fig:2wheels}
\end{figure}

To evaluate the control strategy under a repeatable and physically meaningful disturbance, the rover was intentionally configured with asymmetric wheel traction characteristics. As shown in Fig.~\ref{fig:2wheels}, the rover was equipped with wheels having different grouser heights on the left and right sides: the left wheels employed $15.0\,\mathrm{mm}$ grousers, whereas the right wheels employed $5.4\,\mathrm{mm}$ grousers.

This configuration was motivated by previous terramechanics studies, which demonstrated that grouser geometry strongly influences wheel--soil interaction \cite{takehana2025grouser}. In particular, larger grousers generally increase the traction forces on loose terrain. Consequently, the difference in grouser height produces a persistent imbalance in driving performance between the left and right sides of the rover, resulting in a continuous yaw moment and a systematic heading drift.

By intentionally exploiting the traction asymmetry induced by different grouser heights, the proposed configuration establishes a repeatable experimental benchmark for rover path-following evaluation. Unlike artificial steering disturbances, the benchmark directly originates from wheel--soil interaction, providing a physically interpretable framework for controller assessment on deformable terrain.

\begin{figure*}[t]
    \centering
    \includegraphics[width=1.0\textwidth]{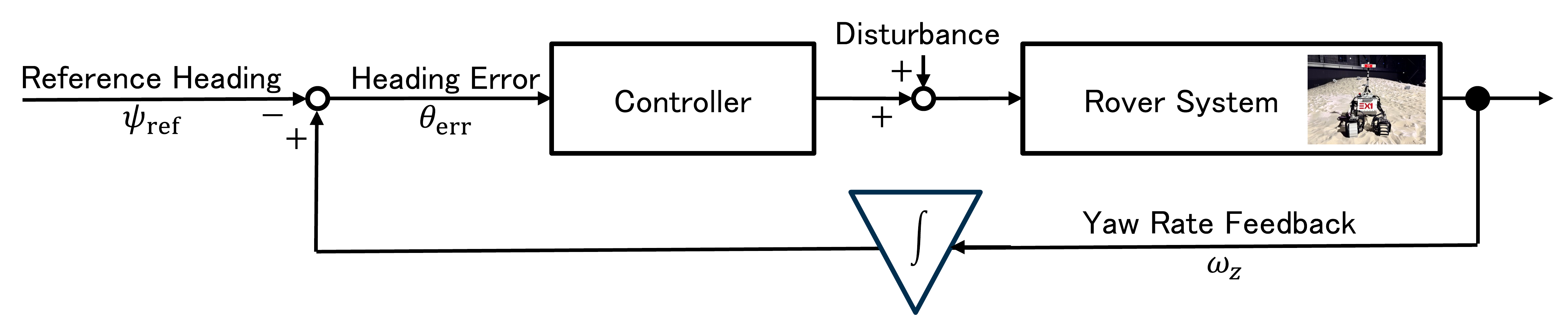}
    \caption{Block diagram of the proposed deceleration-only control system for path following. The corrective yaw moment is generated strictly through deceleration by keeping the maximum commanded velocity at or below $v_{\mathrm{ref}}$.}
    \label{fig:ex1_block}
\end{figure*}

\begin{table}[t]
\caption{Specifications of the experimental rover EX1}
\label{tab:ex1_params}
\begin{center}
\begin{tabular}{lc}
\hline
Parameter & Value \\
\hline
Dimensions (L $\times$ W) & $820\,\mathrm{mm} \times 520\,\mathrm{mm}$ \\
Mass & Approx. $24\,\mathrm{kg}$ \\
Drive system & 4-wheel independent drive \\
Steering mechanism & Skid-steering \\
Maximum speed & $1.0\,\mathrm{m/s}$ \\
Internal sensors & 9-axis IMU, 6-axis force/torque sensors \\
Left wheel grouser height & $15.0\,\mathrm{mm}$ \\
Right wheel grouser height & $5.4\,\mathrm{mm}$ \\
\hline
\end{tabular}
\end{center}
\end{table}

\subsection{Conventional Path Tracking and its Limitations}
In the autonomous navigation of rovers that rely on skid steering, a vehicle typically adjusts its heading by applying a differential in velocity between the left and right wheels while attempting to maintain a nominal reference speed $v_{\mathrm{ref}}$. To generate a corrective yaw moment, conventional controllers that are based on kinematics increase the commanded velocity of the outer wheel and decrease that of the inner wheel. Thus, path correction requires the outer wheel to accelerate beyond the predefined reference speed ($v_{\mathrm{cmd}} > v_{\mathrm{ref}}$).

Although this strategy is effective on rigid surfaces, its application to loose soil inherently introduces challenges in terramechanics. When a conventional controller commands the outer wheel to accelerate beyond the nominal speed on sandy terrain, the accelerated rotation inevitably fails to generate forward thrust due to the shear failure of the soil. Instead, the rotation induces severe slip of the wheel, forcing it to physically excavate the underlying soil, which subsequently leads to the effect of slip-sinkage.

As the wheel sinks into the sand, the resistance to motion increases, thereby significantly degrading the force of traction. This degradation requires additional torque to maintain the commanded speed, which inherently leads to the progressive sinkage of the wheel. Therefore, the conventional requirement to accelerate the outer wheel causes immobilization on soft terrain ($\max(v_{\mathrm{cmd}}) > v_{\mathrm{ref}}$).

The objective of this study is to generate the required corrective yaw moment without causing the cycle of slip-sinkage.

\section{Method}

As illustrated in Fig.~\ref{fig:ex1_block}, the proposed path-following system operates by feeding back the rover's yaw behavior measured by the onboard IMU to calculate the commanded velocities for the left and right wheels. The primary constraint of this system is that the corrective yaw moment is generated through deceleration to prevent the slip-sinkage effect. The control architecture is formulated as follows.

\subsection{Heading Estimation and Error Calculation}
Let $\psi_{\mathrm{ref}}$ be the target heading angle of the rover. When the rover deviates due to terrain variations, the onboard IMU measures the yaw rate $\omega_z$. The actual heading angle $\psi(t)$ is estimated by integrating the measured yaw rate over time:
\begin{equation}
    \psi(t) = \int_{0}^{t} \omega_z(\tau) d\tau
    \label{eq:yaw_angle}
\end{equation}
The heading error $\theta(t)$ is defined as:
\begin{equation}
    \theta(t) = \psi(t) - \psi_{\mathrm{ref}}
    \label{eq:angle_error}
\end{equation}

\subsection{Sensor Noise Mitigation and Proportional Control}
In practice, the measured yaw rate contains noise. To prevent motor chattering and unnecessary soil shear caused by high-frequency components, a deadband constraint is introduced. When the absolute heading error is within a threshold $\theta_{\mathrm{db}}$, the corrective input is set to zero. The filtered heading error $\theta'(t)$ is expressed as:
\begin{equation}
    \theta'(t) = 
    \begin{cases}
        \theta(t) & (|\theta(t)| > \theta_{\mathrm{db}}) \\
        0 & (|\theta(t)| \le \theta_{\mathrm{db}})
    \end{cases}
    \label{eq:deadband}
\end{equation}
The control input $u(t)$ is obtained by multiplying the filtered error by a proportional gain $K_p$:
\begin{equation}
    u(t) = K_p \theta'(t)
    \label{eq:control_input}
\end{equation}

In this implementation, the control loop operated at a frequency of $50.0\,\mathrm{Hz}$. The deadband threshold was set to $\theta_{\mathrm{db}} = 0.04\,\mathrm{rad}$ ($\approx 2.29^{\circ}$), which was selected to exceed the sensor noise floor of the integrated IMU yaw measurement while preventing high-frequency actuator chattering that could induce localized soil shear. The proportional gain was tuned to $K_p = 3.0\,\mathrm{m/(s \cdot rad)}$, providing decisive deceleration authority to correct heading deviations rapidly without destabilizing vehicle yaw. Furthermore, the lower saturation velocity limit was constrained to $v_{\min} = 0.2\,\mathrm{m/s}$ to maintain forward kinetic momentum and prevent the wheels from entering negative rotational speeds that would excavate the regolith.

\subsection{Deceleration-Only Distribution with Saturation Limit}
The control input $u(t)$ is subtracted from the reference velocity to prevent wheel acceleration. To prevent backward wheel rotation under large disturbances, a lower velocity limit $v_{\min}$ is enforced. The final commanded velocities for the left and right wheels ($v_{\mathrm{cmd,L}}$ and $v_{\mathrm{cmd,R}}$) are formulated as:
\begin{equation}
    \begin{cases}
        v_{\mathrm{cmd,L}} = \max \left( v_{\min}, \, v_{\mathrm{ref}} - \max(0, -u(t)) \right) \\
        v_{\mathrm{cmd,R}} = \max \left( v_{\min}, \, v_{\mathrm{ref}} - \max(0, u(t)) \right)
    \end{cases}
    \label{eq:speed_saturation}
\end{equation}

The controller selects the appropriate wheel to decelerate based on the sign of $u(t)$. For example, when the rover drifts to the right ($u(t) < 0$), the controller decelerates only the left wheel to generate a leftward yaw moment. The right wheel maintains the nominal reference speed, ensuring that the maximum commanded velocity never exceeds $v_{\mathrm{ref}}$.

\section{Experiment}

\subsection{Experimental Setup and Methodology}
To evaluate the performance of the proposed deceleration-only path-following control strategy, locomotion tests were conducted in an analogue sandy field at the Advanced Facility for Space Exploration \cite{jaxa2017facility}, JAXA Sagamihara Campus. 

To evaluate the system under terramechanics disturbances, the rover was configured with an asymmetric wheel arrangement; the left wheels were equipped with $15.0\,\mathrm{mm}$ grousers, whereas the right wheels were equipped with $5.4\,\mathrm{mm}$ grousers. This geometric asymmetry generates a thrust imbalance during locomotion, functioning as a steady external disturbance that induces a rightward drift. Open-loop trials (control OFF) and closed-loop trials using the proposed method (control ON) were performed at three reference target velocities ($v_{\mathrm{ref}} = 0.3$, $0.5$, and $0.7\,\mathrm{m/s}$), with three independent trials conducted for each velocity condition.

\begin{figure}[t]
    \centering
    \includegraphics[width=1.0\linewidth]{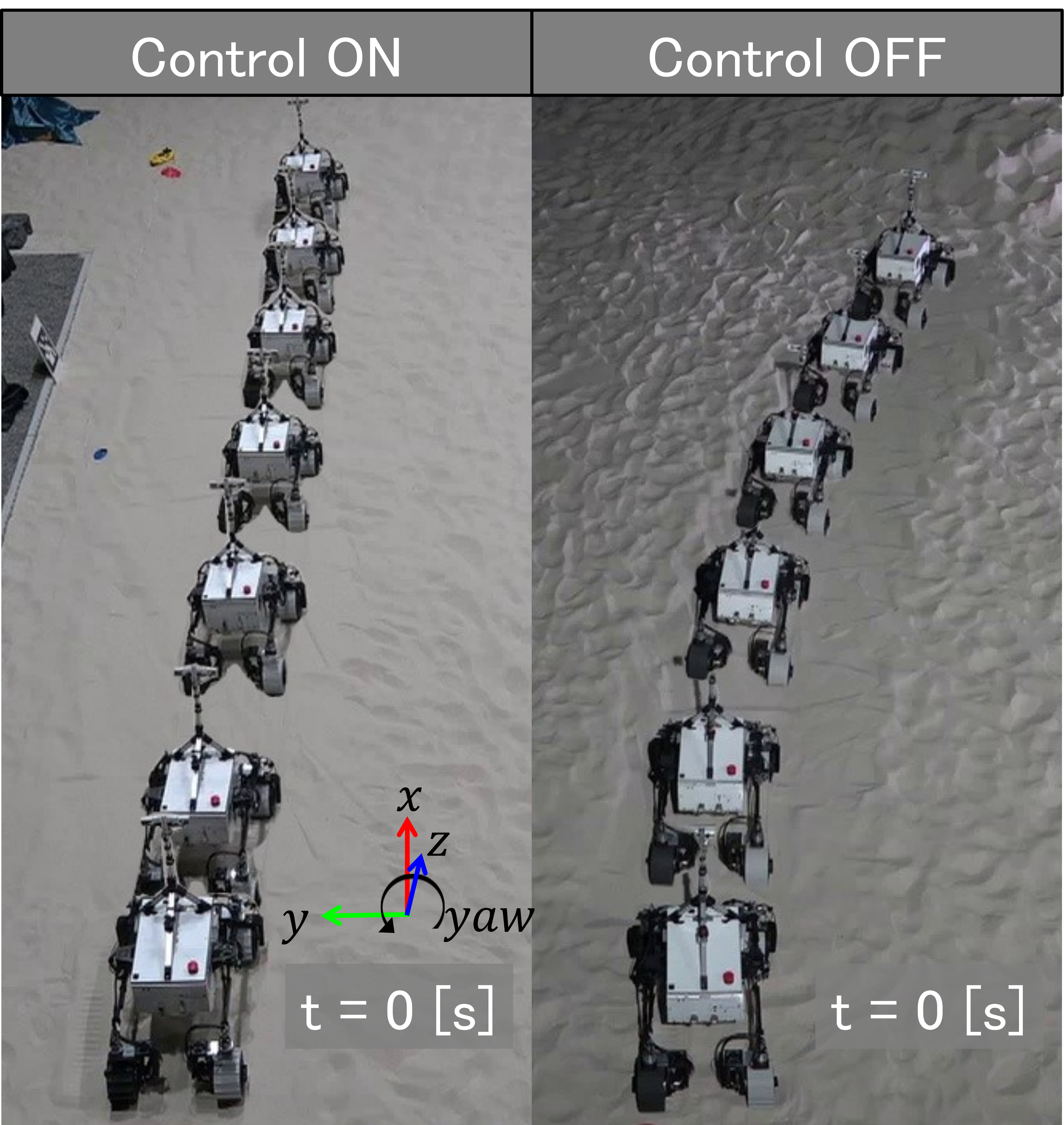}
    \caption{Sequential snapshots of the traversal experiment using the EX1 rover on loose soil under the asymmetric configuration.}
    \label{fig:experiment_snapshots}
\end{figure}

Fig.~\ref{fig:experiment_snapshots} illustrates the process of the traversal experiments under the asymmetric configuration. As these snapshots indicate, the open-loop trial (control OFF) resulted in a steady heading drift. In contrast, the path correction was achieved using the proposed closed-loop control (control ON). To measure the trajectories and lateral deviations, a continuous measurement system was utilized consisting of an automated tracking total station (TS16, Leica Geosystems AG, Heerbrugg, Switzerland) and a 360-degree retroreflective prism. The total station tracked the prism mounted on the rover chassis throughout each run.

\begin{figure*}[t]
    \centering
    \includegraphics[width=1.5\columnwidth]{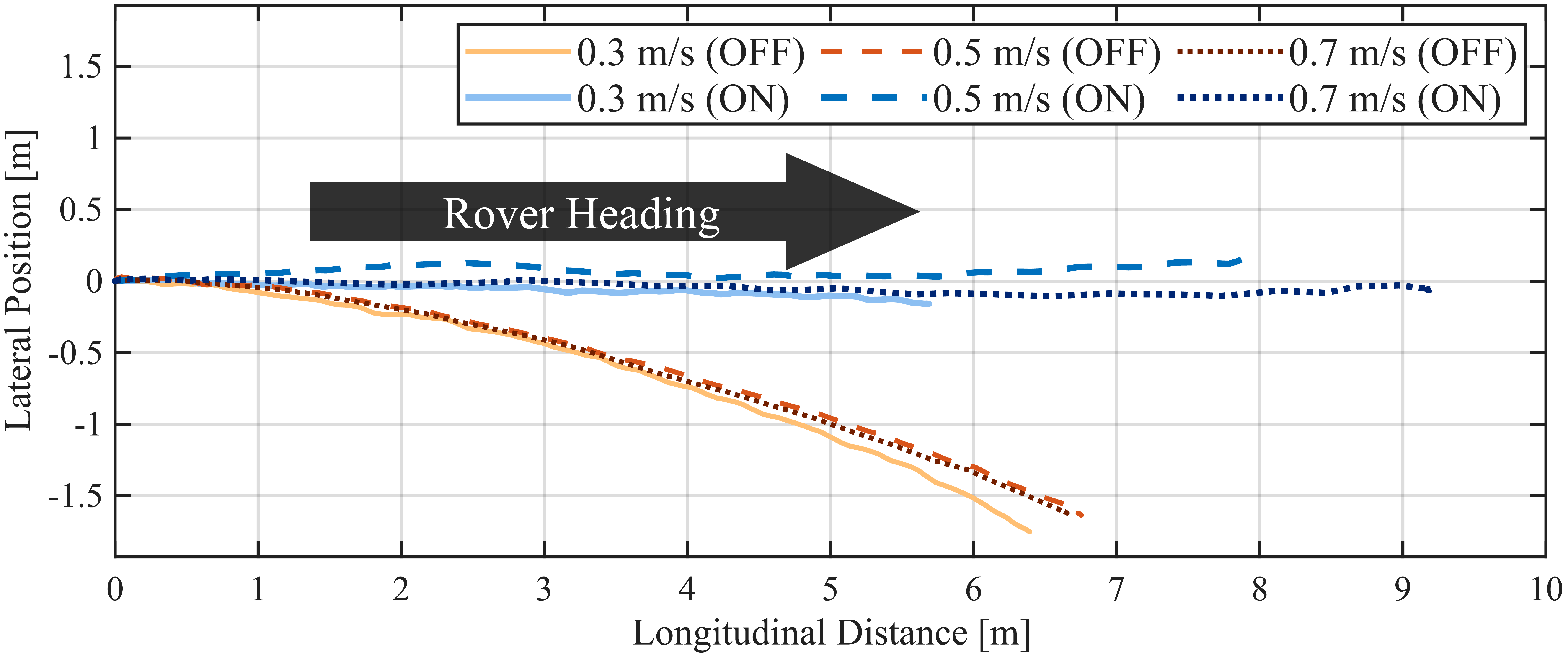}
    \caption{Measured trajectories of the rover with the asymmetric wheel configuration on loose soil at target speeds of $0.3$, $0.5$, and $0.7\,\mathrm{m/s}$. The plotted solid and dashed curves represent the median trajectory filtered from three independent trials.}
    \label{fig:trajectories}
\end{figure*}
    
\begin{figure}[t]
    \centering
    \includegraphics[width=1.0\columnwidth]{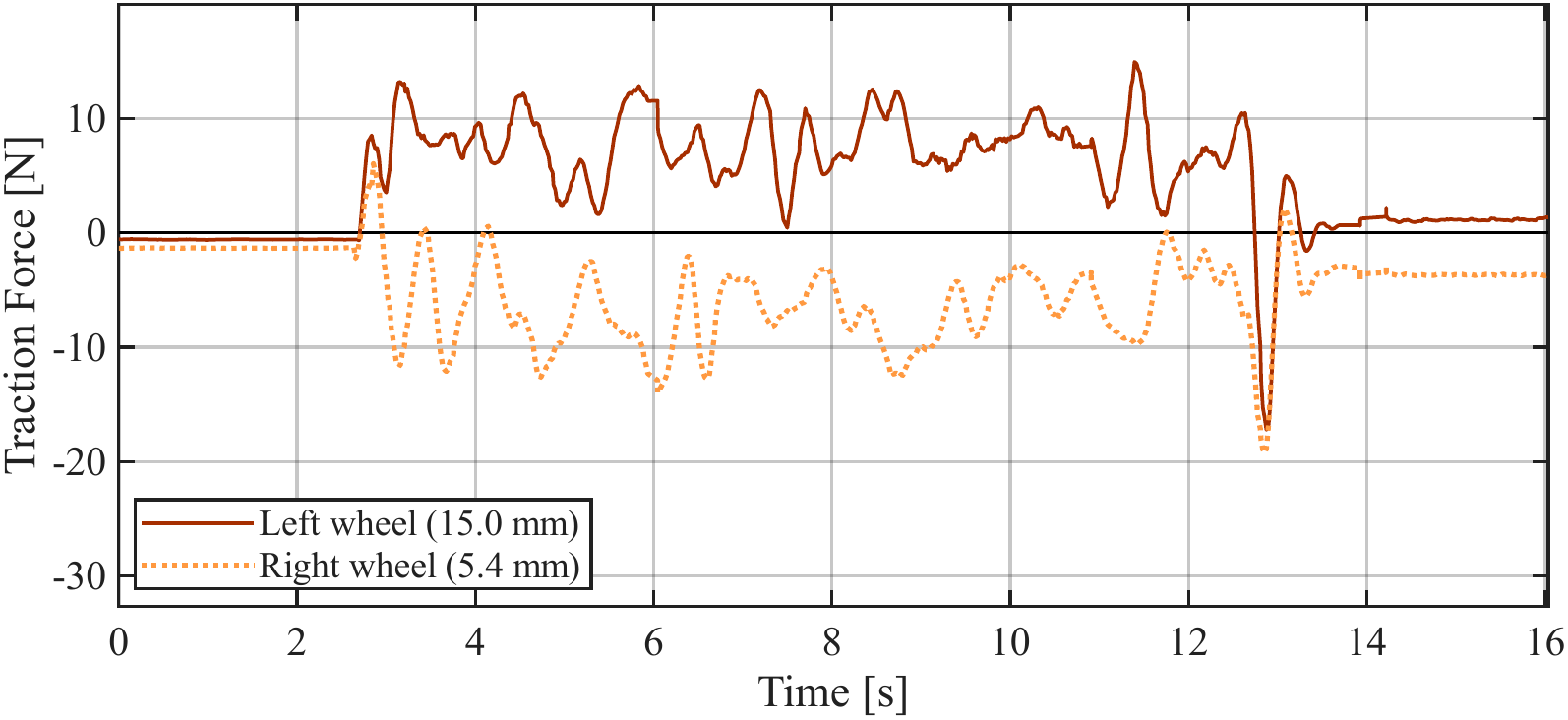}
    \par\vspace{1mm}
    {\footnotesize (a) Open-loop traction forces.}
    
    \vspace{4mm}
    
    \includegraphics[width=1.0\columnwidth]{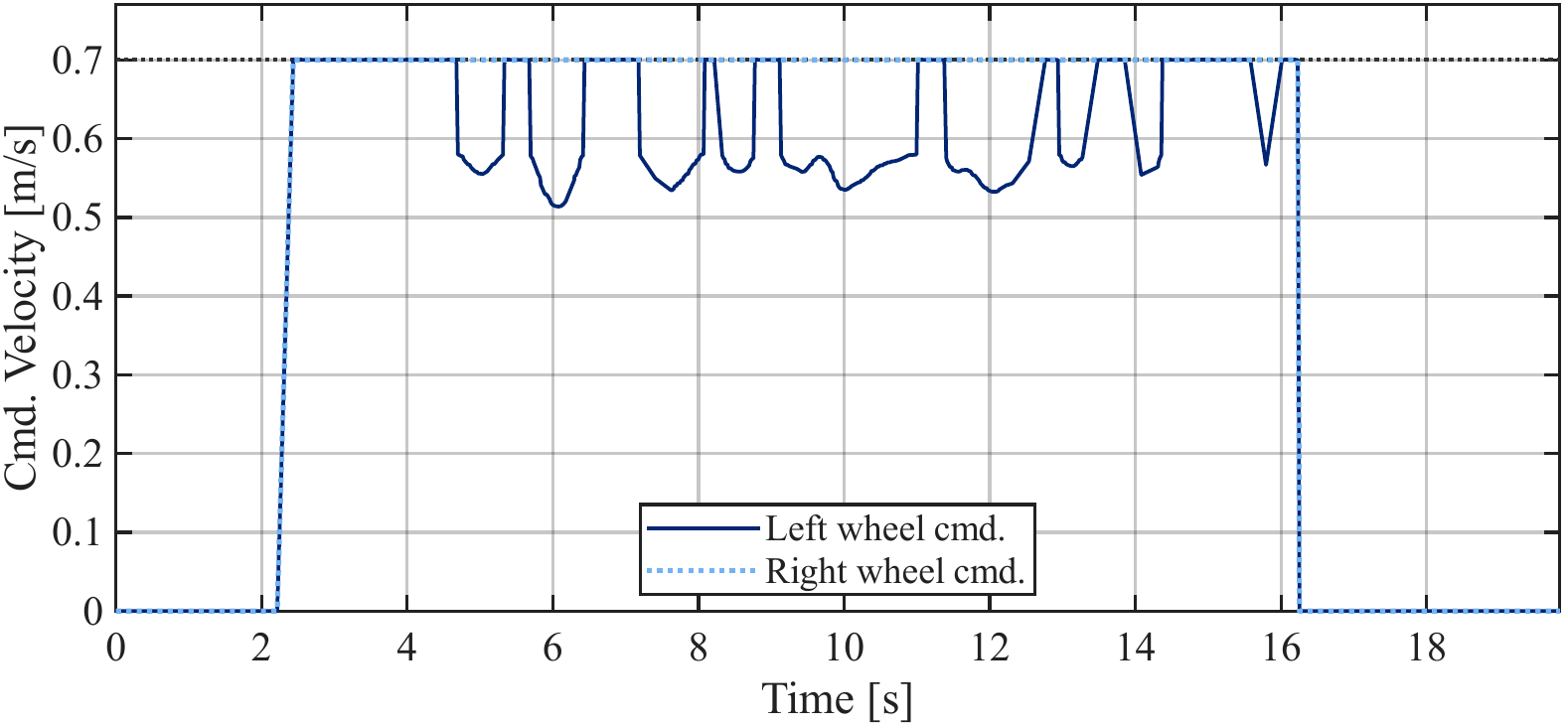}
    \par\vspace{1mm}
    {\footnotesize (b) Command velocities under the proposed control.}
    
    \vspace{4mm}
    
    \includegraphics[width=1.0\columnwidth]{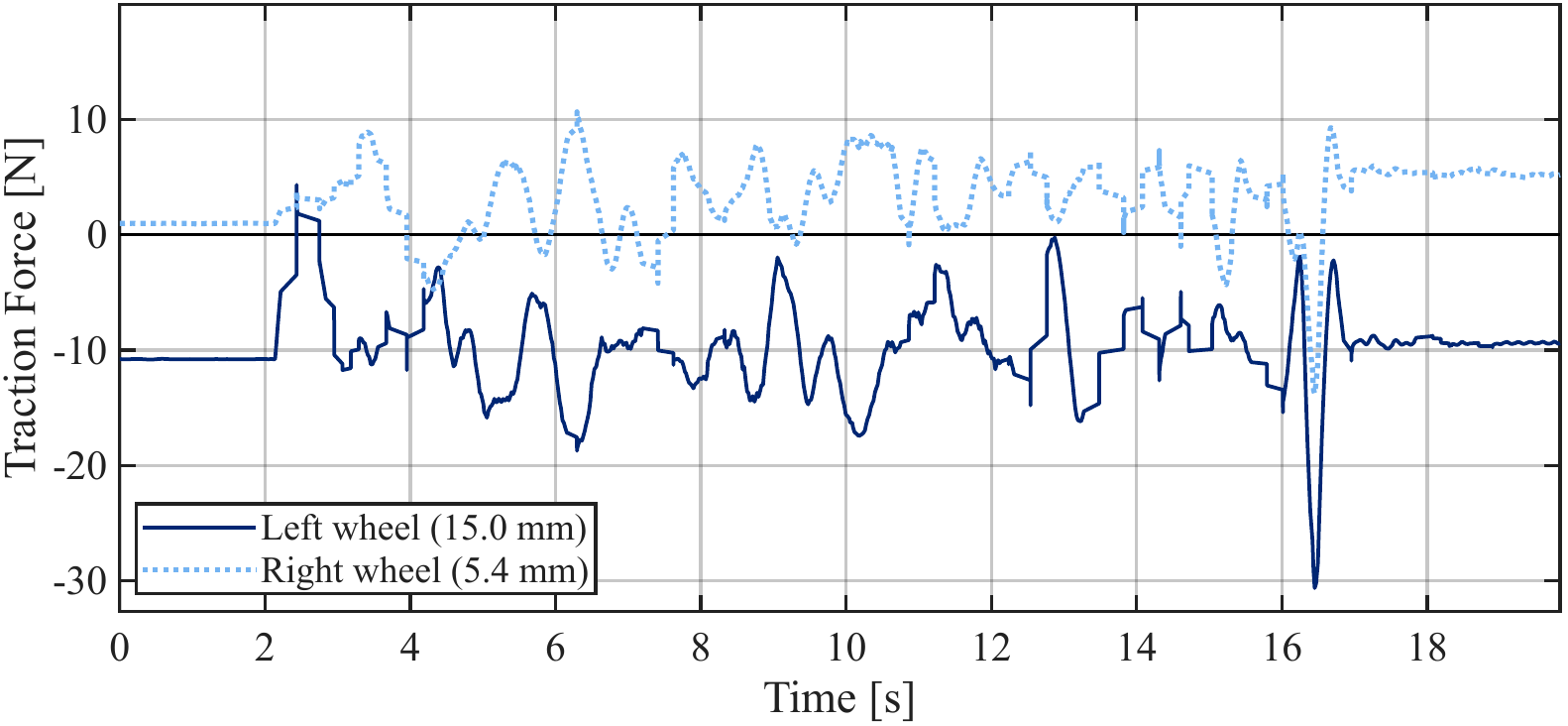}
    \par\vspace{1mm}
    {\footnotesize (c) Traction forces under the proposed control.}
    
    \vspace{2mm}
    \caption{Dynamic response of command velocities and traction forces with asymmetric wheels at $0.7\,\mathrm{m/s}$. The black dotted line in (b) denotes the target nominal speed of $0.7\,\mathrm{m/s}$.}
    \label{fig:dynamics_all_3x1}
\end{figure}

\begin{table*}[h]
\caption{Quantitative Summary of Steady-State Traction Performance Based on All Independent Trials}
\label{tab:traction_stats}
\begin{center}
\begin{tabular}{ccccc}
\hline
\textbf{Target Speed ($v_{\mathrm{ref}}$)} & \textbf{Wheel Component} & \textbf{Control OFF (Open-Loop)} & \textbf{Control ON (Proposed)} & \textbf{Net Improvement ($\Delta$)} \\
\hline
                     & Left Wheel ($F_{x\mathrm{L}}$, $15.0\,\mathrm{mm}$)  & $6.00 \pm 3.35\,\mathrm{N}$ & $-8.22 \pm 7.00\,\mathrm{N}$ & --- \\
$0.3\,\mathrm{m/s}$ & Right Wheel ($F_{x\mathrm{R}}$, $5.4\,\mathrm{mm}$)   & $-5.42 \pm 3.28\,\mathrm{N}$ & $2.71 \pm 4.70\,\mathrm{N}$  & $\mathbf{+8.13\,\mathrm{N}}$ \\
\hline
                     & Left Wheel ($F_{x\mathrm{L}}$, $15.0\,\mathrm{mm}$)  & $7.56 \pm 2.83\,\mathrm{N}$ & $-10.85 \pm 4.60\,\mathrm{N}$ & --- \\
$0.5\,\mathrm{m/s}$ & Right Wheel ($F_{x\mathrm{R}}$, $5.4\,\mathrm{mm}$)   & $-6.04 \pm 2.92\,\mathrm{N}$ & $3.67 \pm 3.82\,\mathrm{N}$  & $\mathbf{+9.71\,\mathrm{N}}$ \\
\hline
                     & Left Wheel ($F_{x\mathrm{L}}$, $15.0\,\mathrm{mm}$)  & $7.54 \pm 2.96\,\mathrm{N}$ & $-9.70 \pm 4.43\,\mathrm{N}$  & --- \\
$0.7\,\mathrm{m/s}$ & Right Wheel ($F_{x\mathrm{R}}$, $5.4\,\mathrm{mm}$)   & $-5.83 \pm 3.79\,\mathrm{N}$ & $3.02 \pm 3.41\,\mathrm{N}$  & $\mathbf{+8.85\,\mathrm{N}}$ \\
\hline
\multicolumn{5}{l}{\footnotesize *Note: Values represent the mean $\pm$ standard deviation pooled from all three independent trials over the active traversal periods.}
\end{tabular}
\end{center}
\end{table*}

\begin{table}[h]
\caption{Quantitative Comparison of Dynamic Normal Loads Showing Load Equalization under Proposed Control}
\label{tab:normal_loads_stats}
\begin{center}
\resizebox{\columnwidth}{!}{
    \begin{tabular}{cccc}
    \hline
    \textbf{Target Speed} & \textbf{Control} & \textbf{Left Wheel $F_z$} & \textbf{Right Wheel $F_z$} \\
    \textbf{($v_{\mathrm{ref}}$)} & \textbf{State} & \textbf{($15.0\,\mathrm{mm}$) [N]} & \textbf{($5.4\,\mathrm{mm}$) [N]} \\
    \hline
                         & OFF & $59.90 \pm 26.21$ & $75.78 \pm 23.81$ \\
    $0.3\,\mathrm{m/s}$ & ON  & $70.20 \pm 24.89$ & $72.50 \pm 27.39$ \\
    \hline
                         & OFF & $46.19 \pm 5.12$  & $64.54 \pm 5.17$ \\
    $0.5\,\mathrm{m/s}$ & ON  & $55.13 \pm 6.82$  & $57.90 \pm 4.86$ \\
    \hline
                         & OFF & $45.73 \pm 6.14$  & $65.20 \pm 6.87$ \\
    $0.7\,\mathrm{m/s}$ & ON  & $54.43 \pm 7.28$  & $57.52 \pm 5.02$ \\
    \hline
    \end{tabular}
}
\end{center}
\end{table}

\subsection{Trajectory Analysis and Path Tracking Performance}
Fig.~\ref{fig:trajectories} illustrates the measured horizontal trajectories of the EX1 rover under the asymmetric configuration. To present a statistically representative behavior for each evaluation scenario, the plotted solid and dashed lines denote the representative trajectory extracted from the independent trials conducted for each velocity regime.

Under open-loop control (dashed lines), the traction imbalance caused by the mismatched grouser heights forced the rover into a substantial rightward drift, causing it to deviate from the desired path within a short longitudinal travel distance. Conversely, upon activating the closed-loop controller (solid lines), the rover exhibited stable path-keeping performance. Despite the continuous external disturbance, the deceleration-only algorithm dynamically stabilized the vehicle's heading and constrained the lateral deviation within a minimal boundary near the reference path ($Y = 0$) across all velocity regimes up to $0.7\,\mathrm{m/s}$ without triggering slip-sinkage immobilization.

\begin{figure}[t]
  \centering
  \subfloat[Left wheel ($F_{x\mathrm{L}}$, $15.0\,\mathrm{mm}$).]{
    \includegraphics[width=0.65\columnwidth]{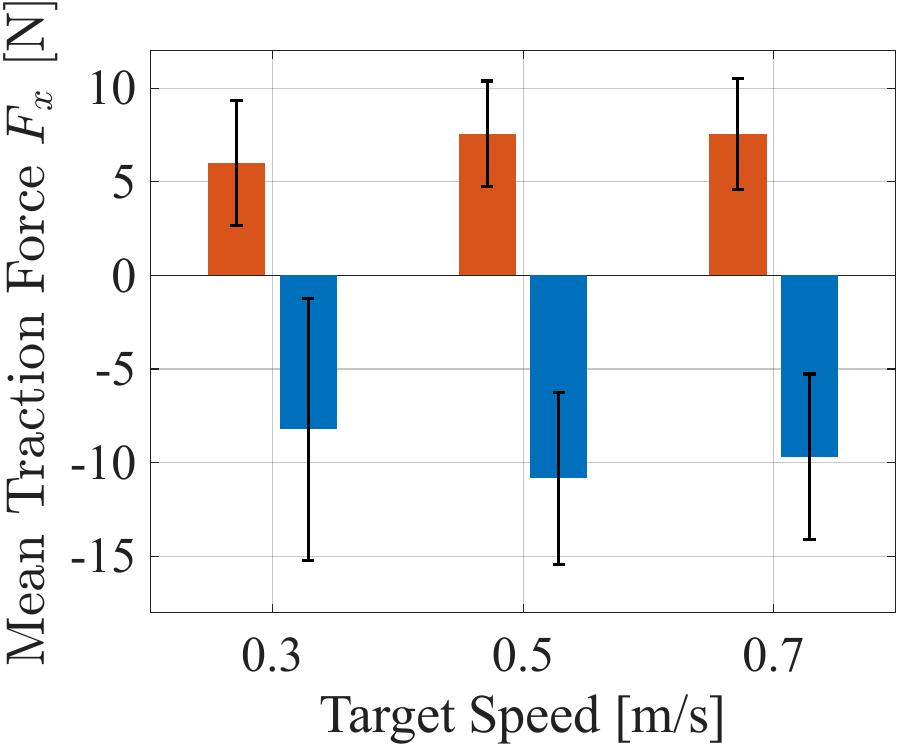}
    \label{fig:bar_traction_left}
  }\\

  \subfloat[Right wheel ($F_{x\mathrm{R}}$, $5.4\,\mathrm{mm}$).]{
    \includegraphics[width=0.65\columnwidth]{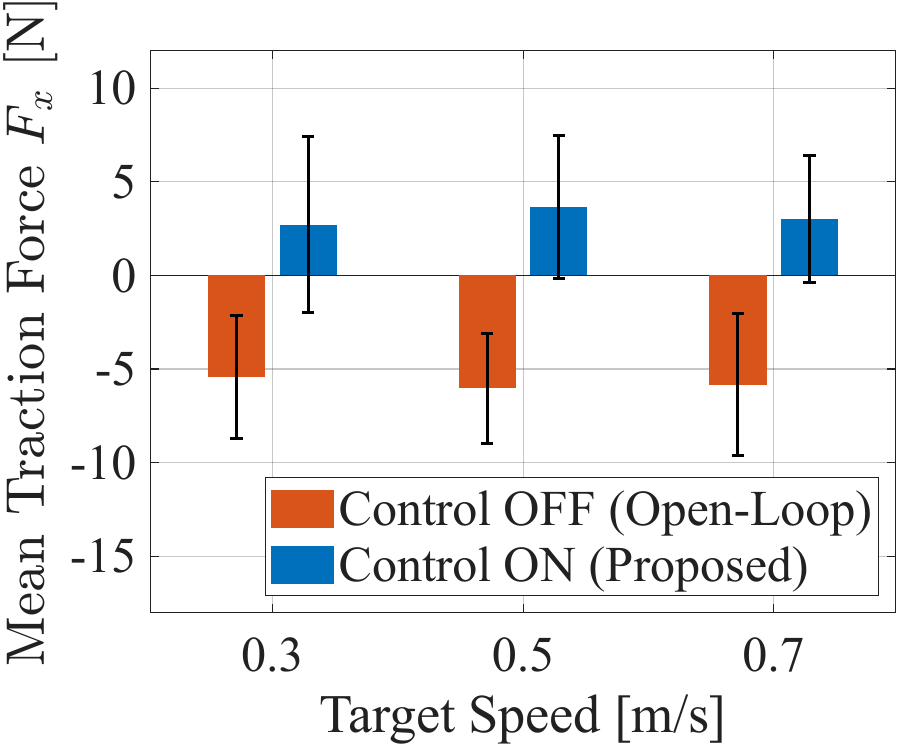}
    \label{fig:bar_traction_right}
  }
  \caption{Quantitative comparison of steady-state mean traction forces under open-loop (OFF) and proposed control (ON) pooled from all independent trials.}
  \label{fig:comprehensive_bar_traction}
\end{figure}

\subsection{Dynamic Traction Analysis and Mechanical Insights}
To elucidate the underlying mechanisms that enable precise path recovery strictly through the deceleration of the outer wheel, Fig.~\ref{fig:dynamics_all_3x1} presents the synchronized data over time regarding the commanded velocities and the forces of bilateral traction ($F_x$), which were recorded during the operation with the asymmetric configuration of the wheels at $v_{\mathrm{ref}} = 0.7\,\mathrm{m/s}$. Furthermore, Table~\ref{tab:traction_stats} provides a quantitative summary of the mean forces of traction pooled from all independent trials, which is visually represented as a chart for comparison in the form of a bar graph in Fig.~\ref{fig:comprehensive_bar_traction}. Additionally, Table~\ref{tab:normal_loads_stats} summarizes the corresponding normal loads ($F_z$) under dynamic conditions. By aggregating and analyzing these combined datasets of forces and states, three fundamental phenomena in mechanics and terramechanics that govern the correction of heading were identified:

\vspace{1mm}

\noindent \textbf{1) Concentration of Loads Induced by Side-Slip and Parasitic Drag (Control OFF):}\\
As quantitatively summarized from the experimental data, during the phase under open-loop control (control OFF), the right wheel, which features shorter grousers ($5.4\,\mathrm{mm}$), consistently fails to generate any positive drawbar pull across various regimes of velocity, dropping into negative values of drag (e.g., $-5.83 \pm 3.79\,\mathrm{N}$ at $0.7\,\mathrm{m/s}$). As clearly shown by the orange bars in the chart for comparison in Fig.~\ref{fig:comprehensive_bar_traction}(b), this wheel is trapped entirely below the line of zero traction. This phenomenon is triggered by a mismatch in kinematics; the left wheel, which generates higher thrust ($7.54 \pm 2.96\,\mathrm{N}$), forces the rover into a continuous drift to the right. Terramechanically, these turning dynamics of skid steering and the large angle of side-slip generate a resistance moment during turning as the vehicle pushes the soil laterally. Consequently, as documented in Table~\ref{tab:normal_loads_stats}, this lateral resistance induces a transfer of lateral loads across the chassis, which inevitably forces a concentration of vertical loads on the right wheel with shorter grousers ($65.20 \pm 6.87\,\mathrm{N}$ compared to $45.73 \pm 6.14\,\mathrm{N}$ for the left wheel). This normal force inherently increases the sinkage and resistance from bulldozing of the wheel with shorter grousers, trapping it in a regime of parasitic drag that subsequently accelerates the deviation from the path.

\vspace{1mm}
\noindent \textbf{2) Symmetric Dissociation of Traction as Structural Evidence of Seesaw Dynamics (Control ON):}\\
Upon activating the proposed control in a closed loop, a distinct transition in the forces and states is captured in the profiles over time (Fig.~\ref{fig:dynamics_all_3x1}(c)). The moment the controller intervenes, the traction of the left wheel sinks deeply into a regime of negative braking ($-9.70 \pm 4.43\,\mathrm{N}$), while the traction of the right wheel simultaneously rises into a regime of positive driving ($+3.02 \pm 3.41\,\mathrm{N}$). Most critically, throughout the entire period of active traversal, the forces of bilateral traction exhibit a synchronized dissociation across a wide range that maintains a striking symmetry in the vertical direction (quasi-symmetric dissociation of bilateral forces). This mirroring profile over time provides experimental evidence regarding the structure of the seesaw effect of the rigid body. Operating through the chassis of the vehicle as a lever arm, the braking force applied electrically to the left side is subsequently converted, one-to-one, into a mechanical torque for forward propulsion that forcefully thrusts the opposite right wheel forward \cite{yoshida2002motion}. 

\vspace{1mm}
\noindent \textbf{3) Path Correction through the Equalization of Loads and Improvement in Net Traction:}\\
As a direct consequence of this seesaw mechanism, which is driven by the generated moment, the heading of the rover is corrected back to the intended straight path ($Y = 0$), which eliminates the unwanted angle of lateral side-slip and the resistance to turning. As a result, the transfer of lateral loads is fully dissipated, which inevitably causes the asymmetric loads in the vertical direction to instantly collapse into a balanced state across the chassis (Left: $54.43 \pm 7.28\,\mathrm{N}$, Right: $57.52 \pm 5.02\,\mathrm{N}$ as listed in Table~\ref{tab:normal_loads_stats}), successfully achieving the equalization of loads. This dynamic mechanism of leverage perfectly explains the improvements in net traction ($\Delta$) of $+8.13\,\mathrm{N}$, $+9.71\,\mathrm{N}$, and $+8.85\,\mathrm{N}$ observed across various regimes of velocity in Table~\ref{tab:traction_stats}, which is visually manifested by the inversion of the traction of the right wheel from the orange bars for the OFF state to the blue bars for the ON state in Fig.~\ref{fig:comprehensive_bar_traction}(b). This net improvement represents a performance of dynamic leverage that significantly exceeds the static limit of the geometry of the shorter grousers; the braking energy of the left wheel is entirely routed through the chassis that acts as a seesaw to lift the right wheel out of its negative regime of drag, simultaneously accomplishing the eradication of parasitic resistance and the complete restoration of the native capability for thrust.

\subsection{Discussion on Kinetic Energy Trade-off and Operating Limitations}
While the deceleration-only strategy successfully suppresses the slip-sinkage cycle on flat terrain, it inherently introduces a physical trade-off regarding the preservation of the rover's forward kinetic energy. Selectively decelerating the outer wheel decreases the net forward drawbar pull during active steering interventions. Under the flat regolith conditions tested in this benchmark, forward momentum was sufficiently maintained by the opposite driving wheel constrained by $v_{\min} = 0.2\,\mathrm{m/s}$. However, in severe operating scenarios---such as ascending steep inclinations or traversing transverse slopes---a persistent reduction in commanded velocity risks depleting forward momentum, which may induce gravitational roll-back or complete traction stall.

Furthermore, while this study focused on a continuous heading disturbance induced by grouser asymmetry, real planetary surfaces feature abrupt, localized terramechanics disturbances (e.g., hidden soft sand pockets or submerged rocks). Against such transient disturbances, the controller immediately activates when heading errors exceed the deadband, utilizing wheel anchoring to resist sudden yaw kicks. Nonetheless, on extreme slopes or high-resistance obstacles where forward momentum is critical, a hybrid control framework that dynamically shifts between deceleration-based anchoring and torque-vectoring will be essential to balance slip prevention against climbability.

\section{Conclusion}
This paper has presented a control strategy for path following that relies on deceleration for planetary rovers that utilize skid steering traversing deformable terrain that is loose. To avoid the cycle of slip-sinkage triggered by the conventional tracking of velocity and acceleration on sandy terrain, the proposed method enforces an upper limit on the maximum velocity that is commanded, correcting deviations in heading solely by selectively decelerating the outer wheels.

The performance of the control and the underlying mechanisms of physics were verified through experiments on locomotion using the EX1 rover under an asymmetric configuration of the wheels that induces disturbances in the heading. Analysis of the trajectories demonstrated that the proposed method suppresses the cumulative drift in the lateral direction across various regimes of velocity up to $0.7\,\mathrm{m/s}$. Furthermore, an analysis of dynamic traction using multi-axis force sensors mounted onboard revealed that the targeted deceleration inherently utilizes resistance from the terrain to act as a mechanical anchor. This effect of anchoring eliminates side-slip, effectively equalizing the dynamic normal loads across the chassis and restoring the opposite wheel that drives to a regime of positive traction. This mechanism of force redistribution, which operates implicitly, demonstrates that prioritizing continuous mobility on soil over strict speed tracking inherently establishes a reliable locomotion strategy for future wide-area, high-speed lunar exploration missions. Future work will extend this framework to complex curved path tracking and evaluate its performance across sloped, heterogeneous terrains with non-uniform soil shear strengths, exploring dynamic velocity baselines to balance slip-sinkage mitigation against kinetic momentum preservation.


\section*{Acknowledgment}
The authors would like to thank the staff of the Advanced Facility for Space Exploration at JAXA Sagamihara campus for their kind support in the field testing.

\bibliographystyle{IEEEtran} 
\bibliography{references}    

\end{document}